\pdfoutput=1
\documentclass[11pt]{article}

\usepackage[final]{acl}

\usepackage{times}
\usepackage{latexsym}
\usepackage{amsmath}
\usepackage{amsfonts} % 提供 \mathbb 命令
\usepackage{float}
\usepackage{graphicx} % 引入图形包
\usepackage{subcaption} % 引入子图包

\usepackage{xcolor}

\usepackage[T1]{fontenc}
\usepackage[utf8]{inputenc}

\usepackage{microtype}

\usepackage{inconsolata}

\usepackage{graphicx}

\title{Step-Wise Formal Verification for LLM-Based Mathematical Problem Solving}

\author{%First Author \\
Kuo Zhou \\ Peking University , 
Beijing , China \\
 \texttt{zhoukuo@pku.edu.cn} \\\And
  Lu Zhang \\ Peking University , 
Beijing , China \\
  \texttt{zhanglu@sei.pku.edu.cn} \\}

\begin{document}
\maketitle
\begin{abstract}
Large Language Models (LLMs) have demonstrated formidable capabilities in solving mathematical problems, yet they may still commit logical reasoning and computational errors during the problem-solving process. Thus, this paper proposes a framework, MATH-VF, which includes a Formalizer and a Critic, for formally verifying the correctness of the solutions generated by large language models.
Our framework first utilizes a Formalizer which employs an LLM to translate a natural language solution into a formal context. Afterward, our Critic (which integrates various external tools such as a Computer Algebra System and an SMT solver) evaluates the correctness of each statement within the formal context, and when a statement is incorrect, our Critic provides corrective feedback.
We empirically investigate the effectiveness of MATH-VF in two scenarios:
1) Verification: MATH-VF is utilized to determine the correctness of a solution to a given problem.
2) Refinement: When MATH-VF identifies errors in the solution generated by an LLM-based solution generator for a given problem, it submits the corrective suggestions proposed by the Critic to the solution generator to regenerate the solution.
We evaluate our framework on widely used mathematical benchmarks: MATH500 and ProcessBench, demonstrating the superiority of our approach over existing approaches.

\end{abstract}

\section{Introduction}
% 第一段 : 概述
Utilizing LLMs for mathematical reasoning is a research area of significant importance, and recent efforts have achieved remarkable progress~\citep{guo2025deepseek,chervonyi2025gold}. 
However, even the most advanced LLMs are still prone to making errors when solving mathematical problems\citep{mirzadeh2024gsm,zhou2024larger,sun2025error}, particularly when the process involves complex logical reasoning and calculations. Therefore, verifying the correctness of solutions generated by LLMs becomes a very important issue.
In addition to verifying the correctness of the solutions, the verifier should also provide feedback on incorrect answers to help the LLM-based generator produce the correct responses.

% 第二段 : 现有研究及其问题
\begin{figure*}[h!]
    \centering
    \vspace{-50pt} % 向上移动10点
    % 第一行
    \begin{subfigure}[b]{0.38\textwidth}
        \centering
        \includegraphics[width=0.8\textwidth, ]{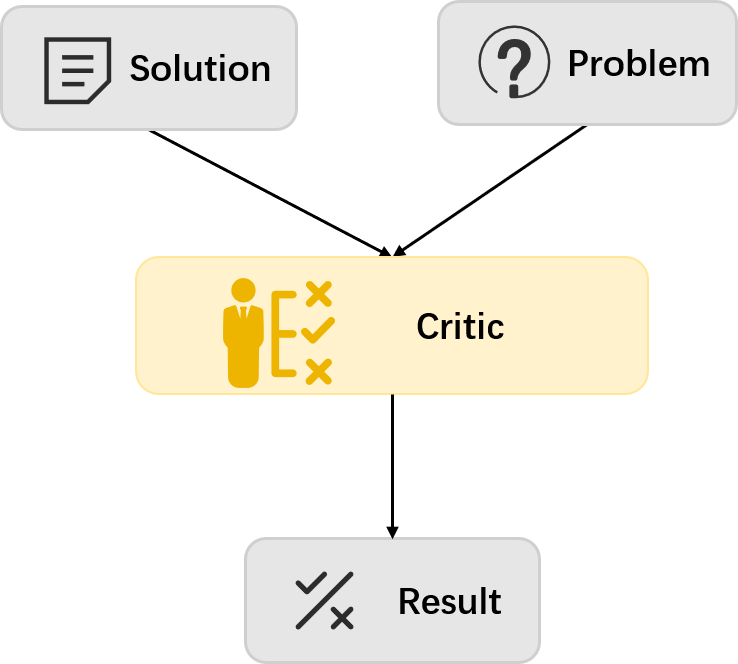}
        \caption{Informal verification}
        \label{fig:sub1}
    \end{subfigure}
    \hspace{1cm} % 添加水平间距
    \begin{subfigure}[b]{0.38\textwidth}
        \centering
        \includegraphics[width=0.8\textwidth]{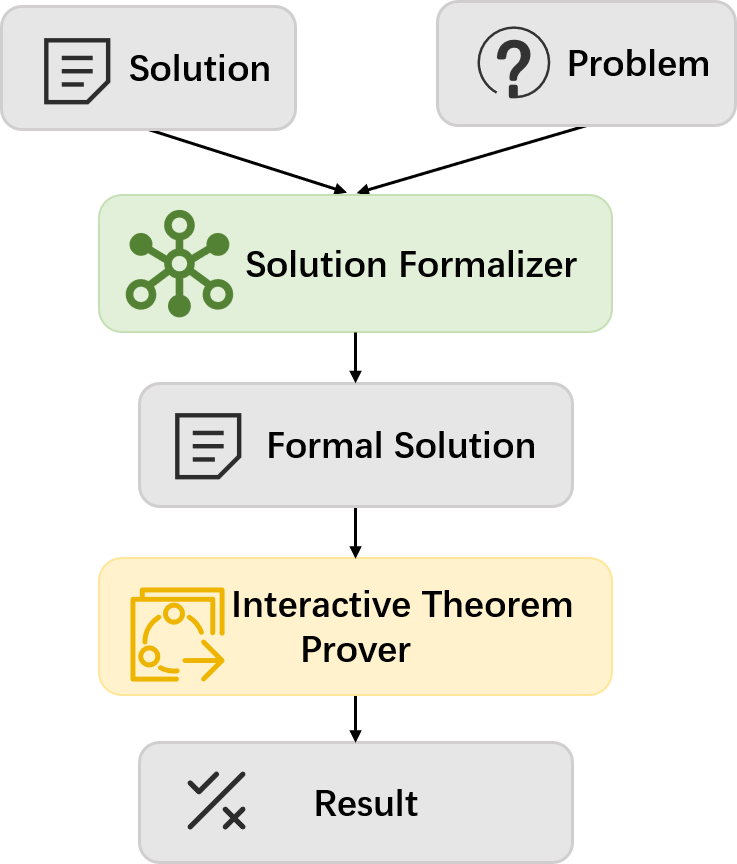}
        \caption{LLM + ITP}
        \label{fig:sub2}
    \end{subfigure}

    \vspace{0.07cm} % 添加1厘米的垂直间距
    % 第二行
    \begin{subfigure}[b]{0.38\textwidth}
        \centering
        \includegraphics[width=0.59\textwidth]{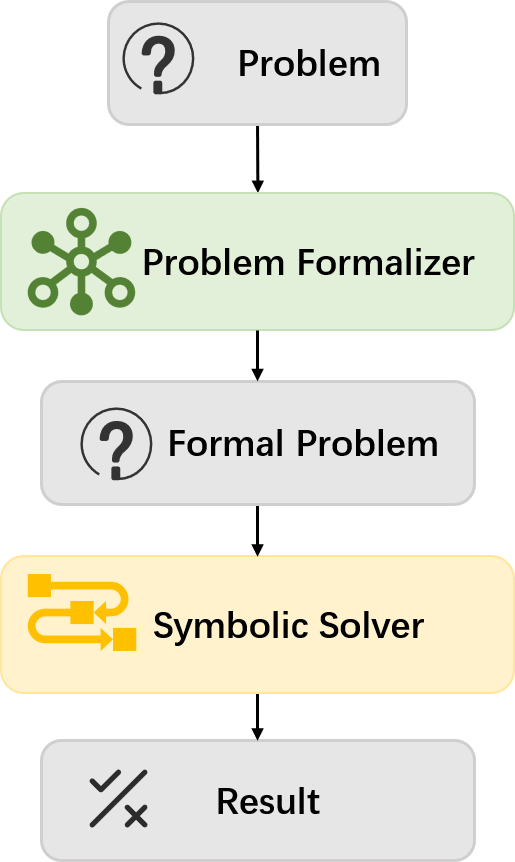}
        \caption{Formalizing problem and solving}
        \label{fig:sub3}
    \end{subfigure}
    \hspace{1cm}  % 添加水平间距
    \begin{subfigure}[b]{0.38\textwidth}
        \centering
        \includegraphics[width=0.8\textwidth]{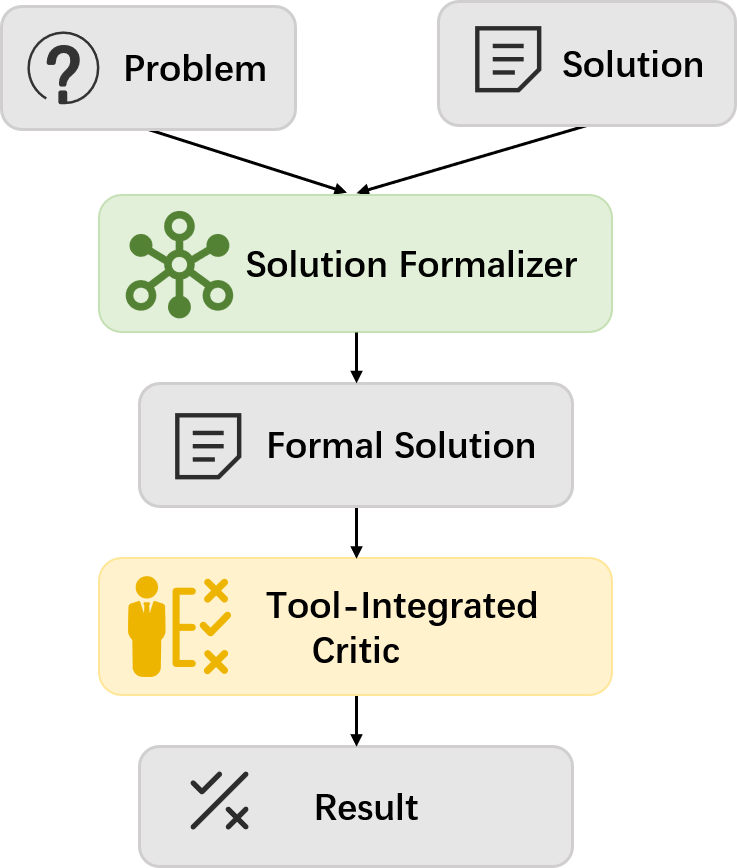}
        \caption{Our method: MATH-VF}
        \label{fig:sub4}
    \end{subfigure}
    \caption{Methods for verifying mathematical reasoning. (a) - (c) from previous work , (d) is our work}
    \label{fig:Four Methods}
\end{figure*}

\iffalse
\begin{figure*}[h!]
    \centering
    % 第一行
    \begin{minipage}[b]{0.38\textwidth}
        \centering
        \includegraphics[width=0.8\textwidth]{latex/method1.png}
        \caption{Informal verification}
        \label{fig:sub1}
    \end{minipage}
    \hspace{1cm} % 添加水平间距
    \begin{minipage}[b]{0.38\textwidth}
        \centering
        \includegraphics[width=0.8\textwidth]{latex/method2.png}
        \caption{LLM + ITP}
        \label{fig:sub2}
    \end{minipage}
    \vspace{0.07cm} % 添加1厘米的垂直间距
    % 第二行
    \begin{minipage}[b]{0.38\textwidth}
        \centering
        \includegraphics[width=0.58\textwidth]{latex/method3.png}
        \caption{Formalizing problem and solving}
        \label{fig:sub3}
    \end{minipage}
    \hspace{1cm} % 添加水平间距
    \begin{minipage}[b]{0.38\textwidth}
        \centering
        \includegraphics[width=0.8\textwidth]{latex/method4.png}
        \caption{MATH - VF}
        \label{fig:sub4}
    \end{minipage}
    \caption{Methods for verifying mathematical reasoning.}
    \label{fig:Four Methods}
\end{figure*}
\fi

In general, as shown in figure~\ref{fig:Four Methods} (\subref{fig:sub1}) - (\subref{fig:sub3}), existing verification methods in this context can be divided into three main categories: 

\noindent\textbf{\textbullet\ Informal verification for natural language reasoning.} 
%2 第一种方法有两类：PRM model, Critic model 
Two types of models have been proposed to verify natural language reasoning: process reward models (PRMs) \citep{lightman2023let,wang-etal-2024-math,khalifa2023grace} and Critic models \citep{luo2023critique,khalifa2023grace}.
A PRM assigns a confidence score to each step of the reasoning process. In contrast, a Critic model provides an evaluation of the correctness of each step in the reasoning process in textual form. However, both PRMs and Critic models cannot avoid the weakness of large language models in handling complex mathematical calculations\citep{lin2024criticbench}. Their assessment of the correctness of each step in the solution remains unreliable. For example, recent studies~\citep{song2025prmbench} have shown that PRMs often struggle to detect fine-grained errors in reasoning processes, and their performance is only slightly better than random guessing. Additionally, models, despite being more powerful than PRMs in some cases, still face challenges in accurately identifying errors, especially when dealing with complex reasoning tasks~\citep{zheng2024processbench}. 

\noindent\textbf{\textbullet\ Formal verification using interactive theorem provers (ITPs).} Approaches in this category typically convert solutions generated by LLMs into a formal language (e.g. Lean\citep{moura2021lean}, Coq\citep{huet1997coq} ,Isabelle\citep{blanchette2011automatic}) and then use interactive theorem provers to verify the formal solutions. Compared to approaches in the first category, approaches in this category are of stronger reliability. However, these approaches still face the following issues: 1) Many solutions expressed in unstructured natural language are difficult to formalize completely~\citep{raza2025instantiation}.
2) Formal proofs generated by the interactive theorem provers can hardly be used as feedback to further guide the LLMs due to the limitations of formal languages in accessibility and usability~\citep{liu2024efficient}.

\noindent\textbf{\textbullet\ Autoformalizing the problem.} 
Recent approaches \citep{pan2023logic,zhoudon,olausson2023linc,ye2024satlm} have focused only on formalizing the problem itself rather than solving it. 
These works use a symbolic solver to solve the formal problem, which brings about issues: 1) When the problem is outside the scope of what the symbolic solver can handle, the solving process is bound to fail. 2) Although these approaches can verify the accuracy of the final answer, they do not detect errors within the intermediate steps of the solving process.
% When we have a problem along with its solution, this approach only allows us to determine whether the final answer is correct, but does not enable us to identify any incorrect steps within the solution process.

\begin{figure*}[h!]
  \includegraphics[width=1\linewidth]{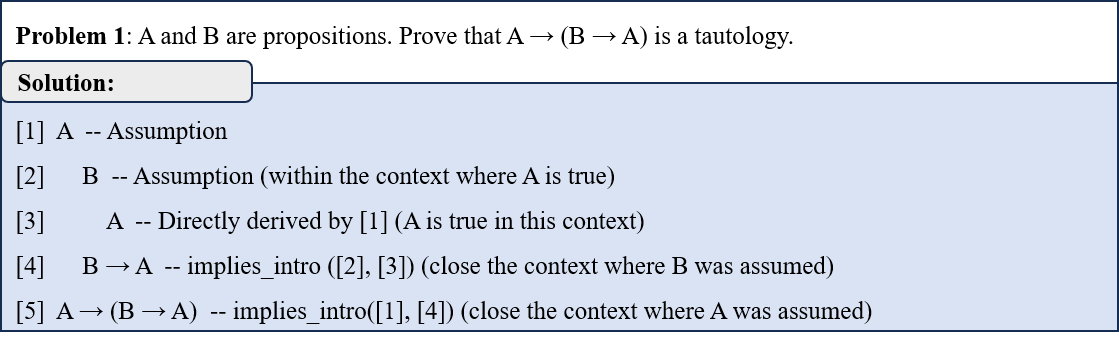}
  \caption{Examle of Fitch-style proof}
  \label{fig:Fitch}
\end{figure*}

To overcome the limitations of existing verification methods, we propose a novel and effective framework, named MATH-VF, for verifying solutions to mathematical problems.
% To tackle the limitations of existing verification approaches, we propose a novel and effective framework for verifying solutions to mathematical problems, named MATH-VF. 
As illustrated in Figure~\ref{fig:Four Methods}(d), 
%\begin{figure}[t]
%  \includegraphics[width=\columnwidth]{example-image-golden}
%  \caption{Overview of MATH-VF}
%  \label{fig:MATH-VF}
%\end{figure}
MATH-VF consists of two main components based on LLMs: the Formalizer and the Critic. 
The Formalizer is prompted to convert natural language solutions into SimpleMath --- a formal language that we have specifically designed as an extension of classical first-order language.The Critic integrates external tools such as SymPy\citep{meurer2017SymPy} and Z3-solver\citep{de2008z3} to enhance its ability to verify the correctness of formal solutions. Figure~\ref{fig:Four Methods} shows the differences between the existing verification methods and MATH-VF. 
In addition, we have observed a phenomenon concerning the relationships among intermediate steps in problem-solving; specifically, some intermediate steps are directly related, while others are not. As illustrated in Figure \ref{fig:Fitch}, Step 4 has direct associations with Step 2 and Step 3, but not with Step 1. Based on this observation, while maintaining the accuracy of judging steps, the complexity of the input information for our Critic has been significantly reduced.
% Despite leveraging external tools to enhance reliability, \red{our Critic in MATH-VF can effectively exploit the sparsity of the solution-graph (see~\ref{subsec:Formalizer}) to reduce the information of the formal context input into the Critic.} This approach significantly reduces computational complexity while preserving verification accuracy.

% This approach significantly lowers computational complexity while maintaining verification accuracy.
% 第四段 : 继续论述

% 第五段 : contribution 总结

\textbf{Our main contributions:} 

1) We propose a formal language (named SimpleMath) based on first-order language, and develop a tool to formalize problem-solving processes expressed in informal languages. The significance of proposing this language, is that the context constructed by first-order language closely resembles the extensive natural language-based mathematical texts that LLMs have learned during their pre-training phase. This makes formalization easier.
% This enhances the conciseness of formalized mathematical representations,
% The reason for proposing this language, is that the context constructed by first-order language closely resembles the extensive natural language-based mathematical texts that LLMs have learned during their pre-training phase， which can reduce the difficulty of formalizing mathematically described problems in natural language.

% We propose a formal language, SimpleMath, based on first-order logic, and develop a tool to formalize problem-solving processes expressed in informal languages. The rationale for this language lies in its ability to construct contexts that closely resemble the natural language-based mathematical texts extensively learned by LLMs during their pre-training phase, thereby facilitating formalization.
% We introduce SimpleMath, a formal language based on first-order logic, and develop a tool to formalize problem-solving processes expressed in informal languages. This language is designed to construct contexts that closely align with the natural language-based mathematical texts extensively learned by LLMs during pre-training, thus simplifying the formalization process."
% This makes formalization easier.

2) We first introduced a critic that integrates the Large Language Model with external tools, such as SymPy and Z3-Solver for determining the correctness of mathematical problem solving steps. Furthermore, we propose a new method to reduce the number of tokens that are entered into our Critic model.

3) Based on Formalizer and Critic, we develop MATH-VF, a training-free framework for step-by-step verification of mathematical reasoning. MATH-VF not only evaluates the correctness of solutions, but also provides constructive feedback for incorrect ones.
% 3) Based on Formalizer and Critic, we develop a train free framework (named MATH-VF) for verifying mathematical reasoning step by step. MATH-VF not only includes the correctness of solutions but also provides constructive feedback for incorrect solutions. 

4) We evaluate our approach on two widely used mathematical benchmarks: MATH500\citep{lightman2023let},ProcessBench\citep{song2025prmbench} ,and our empirical results demonstrate the superiority of our approach over existing approaches.

\section{Related Work}

\subsection{Tool Augmented Language Models}
Early efforts to integrate tools for mathematical reasoning mainly focus on leveraging external calculators, code interpreters\citep{toh2024votescountprogramsverifiers}, and symbolic solvers to address the limitations of traditional language models. For example, MathSensei \citep{das2024mathsensei} incorporates a knowledge retriever (Bing Web Search), and an executor (Python), and a symbolic equation solver (Wolfram-Alpha API) to achieve improved accuracy on complex mathematical reasoning benchmarks. Similarly, the Multi-tool Integration Application framework combines Math Tool, Code Tool, and CoT Tool to perform basic calculations, generate executable code, and enhance logical coherence through iterative reasoning. 
Furthermore, a dataset of interactive tool-use trajectories is created, on which the performance of fine-tuned LLMs is significantly enhanced \citep{gou2023tora}.
% Furthermore, \citep{gou2023tora} create a data set of interactive tool-use trajectories, and fine-tuning a llm on the data set. Through this method, the model performs exceptionally well in mathematical reasoning tasks and significantly enhances its reasoning performance.
In addition to these tools, recent research has also explored the integration of specialized solvers such as Z3 \citep{de2008z3} to handle complex mathematical constraints and symbolic reasoning \citep{pan2023logic}. Z3 is a high-performance theorem prover that can efficiently solve a wide range of mathematical problems, including nonlinear polynomial constraints. By integrating Z3 with language models, the researchers aim to leverage its symbolic reasoning capabilities to improve the overall performance of mathematical reasoning tasks. 
\subsection{Auto Formalization}

There are two types of autoformalization approaches: rule-based approaches and LLM-based approaches. Rule-based approaches \citep{ranta2004grammatical,schaefer2020glif,pathak2024gflean} are deterministic and transparent, making them easier to debug and understand. However, rule-based approaches often struggle with the diversity and complexity of natural languages, leading to limitations in handling edge cases and generalizing to new problem descriptions. 

LLM-based autoformalization leverages large language models (LLMs) to translate mathematical statements from natural languages into formal languages. \citep{wu2022autoformalization} demonstrated that through few-shot learning \citep{wang2020generalizing,parnami2022learning}, LLMs can effectively translate informal mathematical statements into formal specifications in Isabelle/HOL, achieving an accuracy of 25.3\%. 
Other works\citep{xin2024deepseekv1,xin2024deepseekv1_5,azerbayev2023proofnet,} for transforming an informal solution into code that can be verified by interactive theorem provers achieve higher accuracy. However, these works require fine-tuning LLMs on datasets containing a large amount of formalized knowledge and introducing search algorithms, such as BFS and MCTS \citep{browne2012survey,swiechowski2023monte}, in the formalization process. Compared with previous work, the advantage of MATH-VF lies in its ability to achieve high accuracy without fine-tuning.

\subsection{Process Supervision}
Process supervision is designed to evaluate and improve the reasoning capabilities of LLMs by focusing on the intermediate steps of the reasoning process, rather than just the final output. There are two types of Process Supervised Models: Process Reward Models (PRMs) and Critic Models.

\noindent\textbullet\ \textbf{PRMs.} A PRM assigns a score to each individual step in the reasoning process. PRMs are particularly effective in identifying and correcting errors in multi-step mathematical reasoning~\citep{zhang2025lessons}. 

\noindent\textbullet\ \textbf{Critic Models.} The core idea of critic models\citep{kamoi-etal-2024-llms} is to use an LLM as "Critic" to evaluate the correctness of the reasoning process and provide feedback. A step-level Critic dataset MathCritic-76k was proposed to fine-tune a Critic model\citep{xi2024enhancing} . Recent findings by \citep{zheng2024processbench} show that while prompt methods can effectively enable Large Language Models (LLMs) to Critic each solution step by step, existing process reward models typically fail to generalize to more challenging math problems beyond GSM8K and MATH, and underperform compared to Critic models. However, the previous Critic models did not integrate external tools. Therefore, although they can judge the correctness of the solution, their accuracy is limited by the limitations of the computational and reasoning abilities of large models.

\section{Methodology}

\begin{figure*}[h!]
 \centering
  \includegraphics[width=1.06\linewidth]{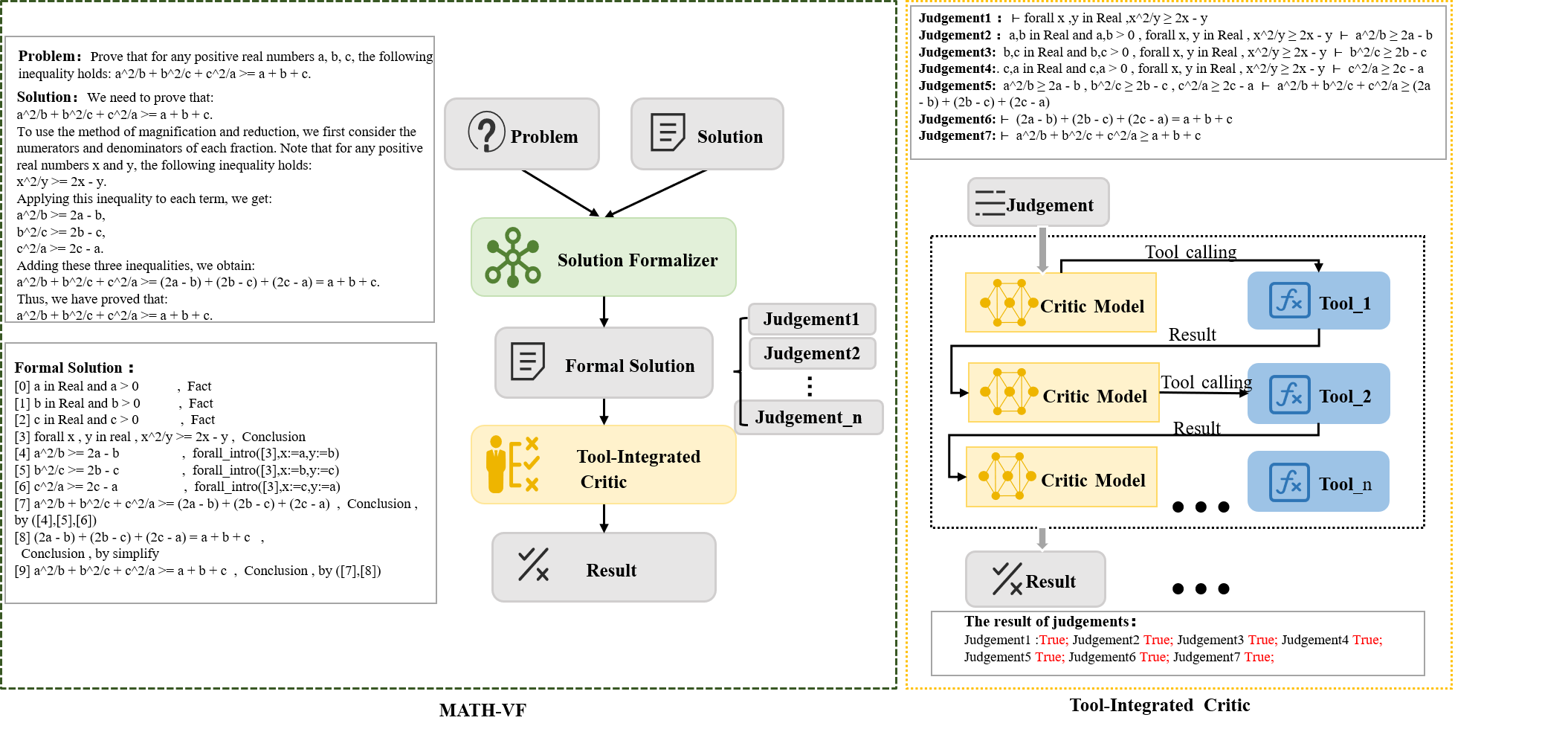}
  \caption{The overview framework of MATH-VF including a solution formalizer and a tool-integreated critic is on the left, and tool-integreated critic is depicted in detail on the right. First, the problem and solution are input into the n solution formalizer, resulting in a context of the formal solution, which can be decomposed into several judgements. And then, we obtain the results through tool-integereated critic which determines the validity of judgments by leveraging both reasoning and tool invocations.}
  % \caption{Overview of MATH-VF: First, the problem and solution are input into the formalizer, resulting in a context of the formal solution. And then We decompose the issue of correctness for the formal solution into a series of correctness issues for individual judgments.}
  \label{fig:framework}
\end{figure*}

As shown in Figure~\ref{fig:framework}, we input the problem and its solution into the formalizer to obtain a context of formal solution. The premises and conclusions in the context may not be continuously derived, with gaps between them. Therefore, we use the Critic to determine whether each conclusion is true under its premises.
% As shown in Figure~\ref{fig:Four Methods}(d), we input the problem and its solution into the formalizer to obtain a context composed of formal language. The premises and conclusions in the context may not be continuously derived, with gaps between them. Therefore, we use the Critic to determine whether each conclusion is true under its premises.

\subsection{Formalizer}\label{subsec:Formalizer} 

% paragraph1:  why formalize ?  and why formalize to first order language

% paragraph2:  How formalize ? -> few shot learning

LLMs have demonstrated a notable ability to comprehend
textual inputs and translate them into formal programs, such as mathematical equations or code. 
We take advantage of the few-shot generalization ability of LLMs to achieve this. By providing LLMs with detailed instructions about the
grammar of the symbolic language and inference rules, together with a few examples in context, we observe that LLMs, such as Deepseek-V3\citep{liu2024deepseek} and GPT-4\citep{achiam2023gpt}, can effectively follow the instructions to translate problems and solutions into a formal context, following our defined grammar and examples.

\begin{figure*}[h]
 
  \includegraphics[width=0.9\linewidth]{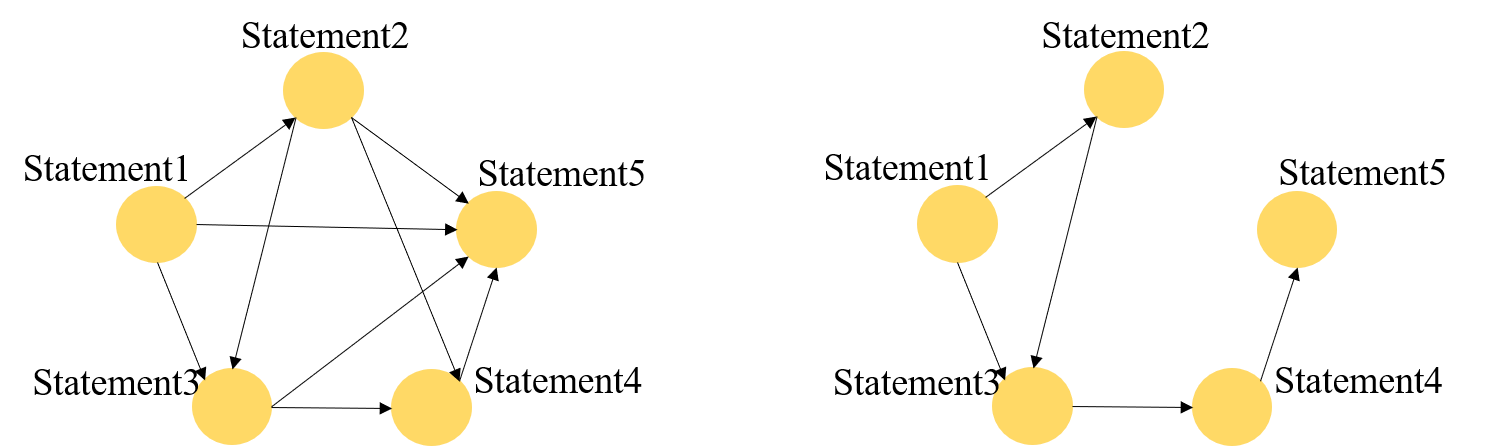}
  \caption{On the left side of the figure is a dense Solution Graph, where each statement is direct conclusion of all previous step's statements. On the right side is a sparse graph, in the solution represented by this graph, statement4 and statement5 have only one premise.}
  \label{fig:sparse}
\end{figure*}

\textbf{SimpleMath Language.}
Our SimpleMath language is an extension of the first-order language, achieved by introducing additional constants and syntactic sugars.
For example :
\begin{align*}
defi&nition(f):\ \mathbb{N} \to \mathbb{N}\\
f(n) &:=\ f(n-1) + f(n-2) , if\ n \geq 3 \ ;\\
  &|\ 1 , if\ n = 2 \ ; \\
  &|\ 1 , if\ n = 1 \ ;
\end{align*}
This definition in SimpleMath has roughly the same effect as the following formular in first order language:
\begin{align*}
        forall\ n &, n\ \in \ \mathbb{N} \to (P_{1}(n) \land P_{2}(n)) \\
        where, & \\
        P_{1}(n) :&\ (n=1\ \lor\ n=2) \to f(n) = 1 \\ 
        P_{2}(n) :&\ (n \geq 3 \to f(n) = f(n-1) + f(n-2)) 
\end{align*}
%forall\ n , n\ \in \ \mathbb{N} \to ((n=1\ \lor\ n=2) \to f(n) = 1 )\\ \land\  (n \geq 3 \to f(n) = f(n-1) + f(n-2))

\textbf{Context.} As illustrated in Figure~\ref{fig:Fitch}, our context is of Fitch Style~\citep{genesereth2022introduction}. In the context, statements can be categorized into five types:

\textbullet\ \textbf{Facts.} A \textbf{Fact} refers to a known condition or piece of information within a problem that is accepted as true without requiring proof. 

\textbullet\ \textbf{Assumptions.} An \textbf{Assumption} refers to a statement that is accepted as true for the purpose of argument, investigation, or problem-solving, even though it may not be proven or verified. 

\textbullet\ \textbf{Theorems.} A \textbf{Theorem} refers to a statement that has been proven to be true based on previously established definitions, facts, and other theorems.

\textbullet\ \textbf{Definitions.} A \textbf{Definitions}refers to a precise statement that clearly explains the meaning of a mathematical term, concept or symbol.

\textbullet\ \textbf{Conclusions.} A \textbf{Conclusion} refers to a statement derived or inferred from known facts, definitions, theorems and previously established conclusions within a problem-solving process. 

Owing to the similarity between SimpleMath and natural language, the accuracy of the formalization results is extremely high. We tested this on the MATH500 dataset, and found that over 90\% of the statements in the natural language solutions generated by the generator can be correctly formalized and verified(sec ~\ref{subsec:result}).

\textbf{Solution Graph\ }

We use Solution Graphs to represent the relationships between different statements within a context, and each context can be transformed into its corresponding Solution Graph.

Given a context, its corresponding Solution Graph is defined as a tuple $(V, E)$, where:

\textbullet\ $V$ is the node set, where each element $v_{i},i= 0, 1 ...  $ corresponds to the the $i$-th statement within the context.

\textbullet\ $E$ is the edge set, where $e_{0,n} , e_{1,n} , ... , e_{n-1,n} \in E$ if and only if $v_{0},...,v_{n} \in V$ and $v_{n}$ is a direct conclusion of $v_{0} , v_{1} , ... , v_{n-1}$.

One of the most significant findings is that Solution Graphs are often sparse (as illustrated in Figure~\ref{fig:sparse}), implying that when a Critic model is verifying the correctness of a particular conclusion, it only needs to input the few statements that are relevant to that conclusion.

%Solution Graph : 求解图 , 利用图的稀疏性来减少推理的复杂度。图的稀疏性就是
Compared to existing methods, our approach effectively leverages the sparsity of the graph, which can significantly reduce the number of input tokens for verification, thereby greatly conserving computational resources. Recent work \citep{ling2024deductive} also observes that LLM can verify each reasoning step by only using irrelevant primise. 
The differences between our method and that work are as follows: 1) Our approach not only extracts information relevant to the conclusion, but also utilizes the relationships among the related information. 
\begin{figure}[h]
  \includegraphics[width=0.8\linewidth]{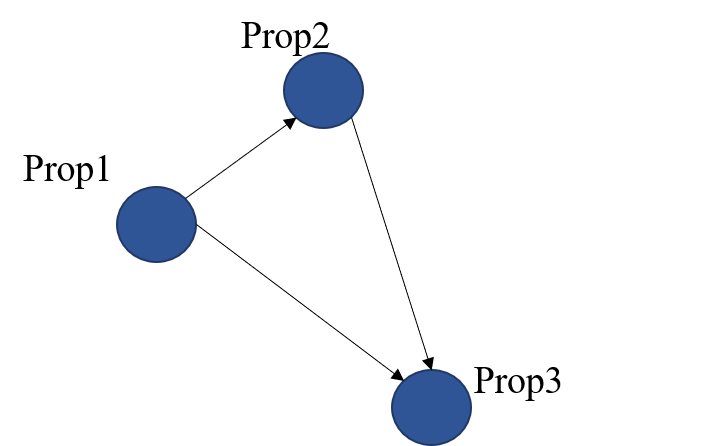}
  \caption{Prop3 is direct conlusion of Prop1 and Prop2 , Prop2 is direct conclusion of Prop2. Therefore we have judgement: Prop1 $\vdash$ Prop3.}
  \label{fig:prop}
\end{figure}
For example, given the Solution Graph as shown in Figure~\ref{fig:prop}, our method can validate a stronger conclusion
$$ \text{Prop}_1 \to \text{Prop}_3, $$
compared to the conclusion
$$ \text{Prop}_1 \land \text{Prop}_2 \to \text{Prop}_3. $$
In contrast, previous work\citep{ling2024deductive}, only considered which premises are relevant to the conclusion, without considering the relationships between premises. So ,the stronger conclusion cannot be validated. We verify the formal context using tools, whereas that work only verifies the context composed of natural language without employing any external tools.

\subsection{Critic}
\iffalse
\begin{figure}[htbp]
  \includegraphics[width=\columnwidth]{latex/critic.png}
  \caption{Overview of our Critic model}
  \label{fig:tool Critic}
\end{figure}
\fi
After the formulator parses the problem $P$ and the solution $S$ into representations $\widehat{P}$ and $\widehat{S} $ ,we obtained a series of judgements: $\mathbb{T}_{1} \vdash Q_{1}, ... ,\mathbb{T}_{i} \vdash Q_{i} $. Here $\mathbb{T}_{i}$ is the context that includes all useful primises, and $Q_{i}$ is the conclusion to verify.  Our Critic model is prompted to evaluate the correctness of these judgments and, when a judgment is erroneous, to provide the reasons for the error.

Our Critic model determines the validity of a judgment by leveraging both reasoning and tool invocations. As shown on the right of Figure~\ref{fig:framework}, given a judgment, our Critic model invokes multiple tools during the reasoning process, working together to verify the judgment.

\subsubsection{Tools for Our Critic Model}
 % Critic判定Judgement有效性的过程构成一个序列 S : r_{1}c_{1}o_{1}...r_{n}c_{n}o_{n} , 这里, r_{i}表示reasoning ,c_{i}表示tool calling.

 % Algorithm 1 表示算法图

% 我们用于判定 Judgement 正确性所需要的 package 主要有三种 1 Computer Algebra System 2 SMT Solver 3 Natural Deduction Prover
\noindent\textbf{Computer Algebra System:} SymPy is a powerful software tool designed to perform both symbolic and numerical computations. It is capable of manipulating mathematical expressions symbolically, performing operations such as differentiation, integration, solving equations, and factoring polynomials. 
% 这里

\noindent\textbf{SMT Solver:} Z3 is a high performance Satisfiability Modulo Theories (SMT) Solver developed by Microsoft Research. It is designed to check the satisfiability of logical expressions and generate models for satisfiable formulas. Z3 supports a wide range of theories, including linear arithmetic (both real and integer), bit vectors, arrays, datatypes, strings, and more.

\subsubsection{Sparsity of the Solution Graph}
As shown in Figure~\ref{fig:sparse}, different Solution Graphs lead to different total input lengths for the Critic to evaluate the solution.
Given a formal context, there are $n$ statements that need to be verified.Defining $C_{1}(n)$ as the total number of statements input to the Critic model in previous work, and $C_{2}(n)$ as the corresponding number in our work. We have:
$$C_{1}(n) = 1 + 2 + 3 + \ldots + n = \frac{n(n + 1)}{2}.$$
and:
$$ C_{2}(n) \leq n \times M.$$ 
, where $M$ is the maximum number of premises of all conclusions. 
According to our statistics, in almost all formal solutions, \ $M \leq 4 $. Thus, in the vast majority of cases, the total number of input statements is less than or equal to $4 \times n$. Compared to other Critic models \citep{xi2024enhancing,zheng2024processbench}, our approach can significantly reduce the number of tokens input into the LLM, especially when the context is long.

\subsection{Prompt Methods for Critic}
We use the few-shot learning approach for our Critic to call tools for reasoning. A small but important distinction from all previous work is that our Critic agent does not directly generate and execute code. Instead, it passes the problem to three LLM-based agents that are integrated with three specific tools. Our experience has shown that using a Critic model as both a reasoner and a tool-agent caller outperforms the use of a single Critic model without any tool-agent.

\subsection{Solution Refinement}
For complex problems, generating the correct reasoning may become a challenge for large language models (LLMs), and it often requires multiple attempts. Moreover, learning from errors is extremely important. Recent work \citep{madaan2024self} proposes a method to improve LLM outputs through iterative feedback and refinement. It uses a single LLM to generate initial responses, provide feedback, and refine them without additional training data. Here, we employ a similar idea: If a solution fails validation, we pass both the solution and the reasons for its failure (provided by our Critic) to the Generator to regenerate the answer. This process repeats until the solution passes validation or reaches the maximum iteration limit.
%For complex problems, generating the correct logical form may become challenging for LLMs. To address this, we introduce a Self-Refinement module that learns to modify inaccurate logical formulations using the error messages from the symbolic reasoner as feedback. Recent works have adopted similar ideas to improve code generation, by teaching LLMs to debug their predicted programs via fewshot demonstrations. Here we extend this idea to refine generated logic representations. If the symbolic solver returns an execution error, we instruct the LLM to refine the incorrect logical form, by prompting it with the erroneous logic form, the solver’s error message, and a set of demonstrations showing common error cases (e.g., a free variable is not bounded to any quantifier in FOL) and their remedies. We run this process iteratively until either no error messages are returned, or the maximum number of allowable revisions is reached

\begin{table}[h!]
\centering
\scalebox{0.60}{% 按照文本宽度调整表格宽度，高度自动调整
\begin{tabular}{lllll}

\hline

\textbf{Generator Model}& \textbf{Acc}. &\textbf{Method} & \ \ \textbf{Discrimin}. \ & \\

\hline

Deepseek - v2.5  & 80.2 &LLM + Coq &\ \  3.9 \  \   &\    \\

\cline{3-5}

                 & &Primary Critic    &\ \  90.6  &\    \\  

\cline{3-5}

                 & &MATH-VF   & \ \ \textbf{93.2}  &\    \\

\hline

Deepseek - v3    & 89.5 & LLM + Coq &\ \ 6.8      &\   \\
\cline{3-5}

             &    & Primary Critic    &\ \  89.5  &\    \\  

\cline{3-5}

                & & MATH-VF&\ \ \textbf{95.7} & \   \\

\hline

Qwen - 2.5 - 72B - Instruct & 82.6 &LLM + Coq & \ \ 4.6  &\    \\  
\cline{3-5}

                 & &Primary Critic    &\ \  87.9  &\    \\  

\cline{3-5}

                 & & MATH-VF&\ \  \textbf{92.4} &  \\

\hline

Qwen - 2.5 - 14B - Instruct & 79.3 & LLM + Coq & \ \ 2.2   & \      \\

\cline{3-5}

                 & &Primary Critic    &\ \  89.3  &\    \\  

\cline{3-5}
                 & & MATH-VF&\ \ \textbf{94.3} &   \\

\hline
\end{tabular}
}
\caption{\label{table : verify}
"Acc". represents accuracy of Generator Model; "Discrimin." refers to the accuracy of verifier in the task of determining whether a reasoning path contains errors
.}
\end{table}

\begin{table}[htbp]
\centering
%\label{tab:performance}
\scalebox{0.60}{% 按照文本宽度调整表格宽度，高度自动调整
\begin{tabular}{l|c|c}
\hline
\textbf{Metric} & \textbf{Qwen2.5-MATH-PRM-72B} & \textbf{MATH-VF} \\
\hline
GSM8K & \textbf{87.3} & 77.2 \\
MATH & \textbf{80.6} & 73.4 \\
OlympiadBench & 74.3 & \textbf{76.1} \\
Omni-MATH & \textbf{71.1} & 69.5 \\
\hline
Avg. & 78.3 & 74.1 \\
\hline
SD. & 0.062& 0.029 \\ 
\hline
\end{tabular}}
\caption{\label{tab:performance} F1 scores of training-required methods and MATH-VF on ProcessBench. %"Avg" represents the average F1 score , while "SD" represents the standard deviation.
}
\end{table}

\iffalse
\begin{table*}[h!]
\centering
\begin{tabular}{|l|c|c|c|c|c|}
\hline
\textbf{model} & \textbf{GSM8K} & \textbf{MATH} 
 &\textbf{OlympiadBench} & \textbf{Omni-MATH} & \textbf{Avg.} \\
\hline
Qwen2.5-Math-PRM-72B & 87.3 & 80.6 & 74.3 & 71.1 & 78.3 \\
\hline
MATH-VF & 77.2 & 73.4 & 76.1 & 69.5 & 74.1 \\
\hline
\end{tabular}
\caption{Performance comparison of different methods on ProcessBenchmark.}
\label{tab:performance}
\end{table*}
\fi
\section{Experiments}
We conducted a series of experiments to compare MATH-VF with existing approaches on these tasks:

\noindent\ \textbf{1)} Determining the correctness of the solution;
\noindent\ \textbf{2)} Identifying the correct solution from the candidates;
\noindent\ \textbf{3)} Refining solutions.

% 1) Simple Math vs. Natural Language vs. Lean

% 2) verifier result 
% 2.1) 作为验证器 
% 2.2) 作为选择器 与 self-con , best of N 比较
% 3) improver 
% 与 no tool Critic 比较

\subsection{Experimental Setup}

\textbf{Dataset:} We evaluate MATH-VF on two benchmaks :  MATH500\citep{hendrycks2021measuring,lightman2023let} 
%\textbf{AIME} \cite{website:AIME} 
and ProcessBench\cite{song2025prmbench}.

\noindent\textbf{Baselines:}

\noindent\textbullet\ For task one, our comparison method is as follows: 1) LLM with Coq: We use the LLM to convert the informal solution into a Coq - formatted context, and then use Coq to verify the correctness of the context. The LLM first formalizes the problem and solution into a theorem to be proved. Then, the LLM formalizes the solution. Finally, Coq is used to verify the formal solution. 2) Primary Critic: We use the LLM to directly evaluate the correctness of the informal solution step by step without tool calling \citep{zheng2024processbench}. 3) Qwen2.5-MATH-PRM-72B\cite{zhang2025lessons}.

\iffalse
\begin{table*}[h!]
\centering
\begin{tabular}{lllll}

\hline

\textbf{Generator Model}& \textbf{Acc}. &\textbf{Method} & \ \ \textbf{Discrimin}. \ & \\

\hline

Deepseek - v2.5  & 80.2 &LLM + Coq &\ \  3.9 \  \   &\    \\

\cline{3-5}

                 & &Primary Critic    &\ \  90.6  &\    \\  

\cline{3-5}

                 & &MATH-VF   & \ \ \textbf{93.2}  &\    \\

\hline

Deepseek - v3    & 89.5 & LLM + Coq &\ \ 6.8      &\   \\
\cline{3-5}

             &    & Primary Critic    &\ \  89.5  &\    \\  

\cline{3-5}

                & & MATH-VF&\ \ \textbf{95.7} & \   \\

\hline

Qwen - 2.5 - 72B - Instruct & 82.6 &LLM + Coq & \ \ 4.6  &\    \\  
\cline{3-5}

                 & &Primary Critic    &\ \  87.9  &\    \\  

\cline{3-5}

                 & & MATH-VF&\ \  \textbf{92.4} &  \\

\hline

Qwen - 2.5 - 14B - Instruct & 79.3 & LLM + Coq & \ \ 2.2   & \      \\

\cline{3-5}

                 & &Primary Critic    &\ \  89.3  &\    \\  

\cline{3-5}
                 & & MATH-VF&\ \ \textbf{94.3} &   \\

\hline
\end{tabular}
\caption{\label{table : verify}
"Acc". represents accuracy of Generator Model; "Discrimin." refers to the accuracy of verifier in the task of determining whether a reasoning path contains errors
.}
\end{table*}
\fi

\noindent\textbullet\ For task two, we compare MATH-VF to Self-consistency \citep{wangself} and Primary Critic.

\noindent\textbullet\ For task three, we compare MATH-VF with the self-refinement method \citep{madaan2024self}. In Self-Refinement, we do not need a Critic; we only require the generator to iteratively produce solutions and conduct self-evaluation.\\
For all tasks, we use DeepSeek-v3 as Formalizer and Critic.
\subsection{Main Results}\label{subsec:result}

\textbf{Task one: Determining the correctness of the solution}
\iffalse

\begin{table}
\centering
\begin{tabular}{lllll}
\hline
\textbf{Generator Model}& \textbf{MATH500} \ & \textbf{AIME} \\

\hline
Deepseek-v2.5  &   80.2 & 16.7 \\
\hline

Deepseek-v3  &    88.5 & 38.1  \\
\hline

Qwen-2.5-72B-Instruct  &    82.6 &  15.9 \\
\hline

Qwen-2.5-14B-Instruct &    78.2 &  10.2 \\
\hline
\end{tabular}
\caption{\label{table : generator correctness}
Citation commands supported by the style file.
The style is based on the natbib package and supports all natbib citation commands.
It also supports commands defined in previous ACL style files for compatibility.
}
\end{table}

\fi
We report the results of MATH-VF and the baselines in Table ~\ref{table : verify}. We have the following observations:
1) Given a correct informal solution, generating a formal solution that can be verified by Coq is extremely challenging. In our task-one experiment, among all the correct informal solutions generated by LLM, however, less than 10\% of their corresponding formal solutions can pass the Coq verification. 
This indicates that generating Coq code poses a significant challenge for LLMs, if we only use in-context learning.

2) For the solutions generated by different models, the success rate of MATH-VF in determining the correctness of the solutions is similar. This indicates that MATH-VF can be used to assess the correctness of the solutions generated by various models. 

A formal proof in Coq must include all the details of the reasoning to pass the check of interactive theorem provers. However, in MATH-VF, gaps between the conclusion and the premises are allowed, making formalization in MATH-VF relatively easier. Moreover, such gaps are often easily fillable. 
Another key point is that, compared to Coq, which is based on dependent type theory, SimpleMath is closer to the informal mathematical language that LLMs have learned during their pre-training phase, and this makes the formalization process more straightforward.

The questions in ProcessBench are sourced from GSM8K, Math, OlympiadBench, and Omni-MATH. Among these, GSM8K and Math are relatively simple, while OlympiadBench and Omni-MATH are more difficult.As shown in table ~\ref{tab:performance}, our approach has a lower average F1 score than training-required method : Qwen2.5-MATH-PRM-72B. However The correctness of our approach is more stable and does not show significant decay as the difficulty of the questions increases. We believe the reason that our method is more stable is that, during the formalization process, we break down the answers into finer-grained derivations. Therefore, the difficulty of judging each step in the problem-solving process does not increase with the overall difficulty of the problem. Qwen-PRM-72B requires training on a large amount of data, while our method is training-free. This means that the application scenarios of these two methods are not entirely the same. For example, in our method, we can use closed-source models as formalizers and critics. However, it is infeasible to use data to train closed-source models to serve as PRMs.

\textbf{Task two: identifying the correct solution from the candidates.}
\begin{table}
\centering
\scalebox{0.78}{% 按照文本宽度调整表格宽度，高度自动调整
\begin{tabular}{lllll}

\hline

\textbf{Generator Model}&\textbf{Method} &  \textbf{Acc}. \ & \\

\hline

Deepseek - v2.5  & Self-Consistency & 83.1  &\  \\

\cline{2-5}
                & Primary Critic &  82.9  &\  \\
\cline{2-5}

 & MATH-VF& \textbf{83.6} &  &  \\

\hline

Deepseek - v3  & Self-Consistency &  89.9  &\   \\

\cline{2-5}

            &Primary Critic &  90.3  &\   \\

\cline{2-5}

 & MATH-VF& \textbf{90.5} &   & \\

\hline

Qwen - 2.5 - 72B - Instruct & Self-Consistency & \textbf{87.1}   &\   \\

\cline{2-5}
                &Primary Critic & 84.5  &\  \\

\cline{2-5}

 & MATH-VF& 86.4 &   &\\

\hline

Qwen - 2.5 - 14B - Instruct & Self-Consistency &  83.7   &\   \\

\cline{2-5}

                &Primary Critic & 87.6  &\  \\
\cline{2-5}

 & MATH-VF& \textbf{88.7} &   &\\

\hline
\end{tabular}
}
\caption{\label{table : select}
Performance of MATH-VF , Self-Constency ,and Primary Critic. "Acc." represents the proportion of the identified solutions that are correct.
}

\end{table}
We generate eight candidate solutions for each problem. For self-consistency, we select the answer with the highest consistency score among the candidate answers as the final answer. For MATH-VF, we choose the solution in which every step is evaluated as correct as the final answer. If more than one solution is evaluated as correct, we select the one with the fewest number of statements. The primary Critic follows the same steps as MATH-VF. Table~\ref{table : select} presents that compared to the other two methods, the solutions identified by MATH-VF are more likely to be correct.
When the generator is relatively weak, the identification process leads to more significant improvements.

\textbf{Task three: Refine solutions.} For task three, we use two methods for refinement. As shown in Table~\ref{table : refine}: Self-Refinement contributes very little to the improvement of accuracy.
The main reason MATH-VF outperforms Self-Refinement is that MATH-VF utilizes external tools to offer more accurate suggestions for improvement. Although in the Self-Refinement approach, LLMs evaluate solutions after generation to improve accuracy, it still struggles to overcome the inherent limitations of LLMs. 

\begin{table}
\centering
\scalebox{0.78}{
\begin{tabular}{lllll}

\hline

\textbf{Generator Model}&\textbf{Method} &  \textbf{Acc}. \ &\  \\

\hline

Deepseek - v2.5  & Self-Refine & 80.9  &\  \\

\cline{2-5}

 & MATH-VF& \textbf{83.5} &  &  \\

\hline

Deepseek - v3  & Self-Refine &  90.1  &\   \\

\cline{2-5}

 & MATH-VF& \textbf{92.4} &   & \\

\hline

Qwen - 2.5 - 72B - Instruct & Self-Refine &  83.7   &\   \\

\cline{2-5}

 & MATH-VF& \textbf{86.2} &   &\\

\hline

Qwen - 2.5 - 14B - Instruct & Self-Refine &  80.1   &\   \\

\cline{2-5}

 & MATH-VF& \textbf{82.6} &   &\\

\hline
\end{tabular}
}
\caption{\label{table : refine}
Performance of MATH-VF and Self-Refine.
}
\end{table}

%\subsection{Case Study}

\section{Conclusion}
In this work, we propose a novel step-by-step approach for verifying solutions of mathematical problems. In our approach: MATH-VF, the Formalizer first formalizes the informal solution, and then the Critic leverages external tools such as Z3 and SymPy to verify the conclusions in each step of the reasoning. We evaluated MATH-VF on the MATH500 and ProcessBench, showed that :1) compared to existing training-free methods, MATH-VF performs better in verification, identification, and refinement.
2) compared to existing training-required methods, Math-VF shows inferior accuracy, yet demonstrates greater stability.
Compared to previous work, the main advantages of MATH-VF are: 1) unlike PRM, our method is training-free. This makes our approach more compatible with closed-source models. 2) our approach reducing the number of premises input into the LLM during the verification process.
3) our approach  exhibits stronger stability when dealing with problems of varying difficulty.
\section*{Limitations}

Our paper has some limitations, which we leave for
future work:


First, when Formalizer converts natural language into formal language, the expressions may contain syntax errors, leading to the failure of Critic when calling the solver. Future works could explore the development of using a syntax parser to parse the expressions.

Second, MATH-VF can only assess the correctness of each step in a solution without analyzing the effectiveness of each step (i.e., whether this step brings us closer to the goal). Future research could explore the study of the effectiveness of each step, thereby enhancing MATH-VF's performance in the solution refinement task. 

% \section*{Acknowledgements}
% This paper is supported by xxxxxxx of China xxxxxxxxxxxxx.
% This paper is supported by \red{the National Key Research and Development Program of China 2020AAA0106700}.

% This work was supported in part by the Shenzhen Science and Technology Program under Grant No. JCYJ20200109113201726 and the National Natural Science Foundation of China under Grant No.61872108.

\bibliography{custom}

\begin{thebibliography}{47}
\providecommand{\natexlab}[1]{#1}

\bibitem[{Achiam et~al.(2023)Achiam, Adler, Agarwal, Ahmad, Akkaya, Aleman, Almeida, Altenschmidt, Altman, Anadkat et~al.}]{achiam2023gpt}
Josh Achiam, Steven Adler, Sandhini Agarwal, Lama Ahmad, Ilge Akkaya, Florencia~Leoni Aleman, Diogo Almeida, Janko Altenschmidt, Sam Altman, Shyamal Anadkat, et~al. 2023.
\newblock Gpt-4 technical report.
\newblock \emph{arXiv preprint arXiv:2303.08774}.

\bibitem[{Azerbayev et~al.(2023)Azerbayev, Piotrowski, Schoelkopf, Ayers, Radev, and Avigad}]{azerbayev2023proofnet}
Zhangir Azerbayev, Bartosz Piotrowski, Hailey Schoelkopf, Edward~W Ayers, Dragomir Radev, and Jeremy Avigad. 2023.
\newblock Proofnet: Autoformalizing and formally proving undergraduate-level mathematics.
\newblock \emph{arXiv preprint arXiv:2302.12433}.

\bibitem[{Blanchette et~al.(2011)Blanchette, Bulwahn, and Nipkow}]{blanchette2011automatic}
Jasmin~Christian Blanchette, Lukas Bulwahn, and Tobias Nipkow. 2011.
\newblock Automatic proof and disproof in isabelle/hol.
\newblock In \emph{Frontiers of Combining Systems: 8th International Symposium, FroCoS 2011, Saarbr{\"u}cken, Germany, October 5-7, 2011. Proceedings 8}, pages 12--27. Springer.

\bibitem[{Browne et~al.(2012)Browne, Powley, Whitehouse, Lucas, Cowling, Rohlfshagen, Tavener, Perez, Samothrakis, and Colton}]{browne2012survey}
Cameron~B Browne, Edward Powley, Daniel Whitehouse, Simon~M Lucas, Peter~I Cowling, Philipp Rohlfshagen, Stephen Tavener, Diego Perez, Spyridon Samothrakis, and Simon Colton. 2012.
\newblock A survey of monte carlo tree search methods.
\newblock \emph{IEEE Transactions on Computational Intelligence and AI in games}, 4(1):1--43.

\bibitem[{Chervonyi et~al.(2025)Chervonyi, Trinh, Ol{\v{s}}{\'a}k, Yang, Nguyen, Menegali, Jung, Verma, Le, and Luong}]{chervonyi2025gold}
Yuri Chervonyi, Trieu~H Trinh, Miroslav Ol{\v{s}}{\'a}k, Xiaomeng Yang, Hoang Nguyen, Marcelo Menegali, Junehyuk Jung, Vikas Verma, Quoc~V Le, and Thang Luong. 2025.
\newblock Gold-medalist performance in solving olympiad geometry with alphageometry2.
\newblock \emph{arXiv preprint arXiv:2502.03544}.

\bibitem[{Das et~al.(2024)Das, Banerjee, Aditya, and Kulkarni}]{das2024mathsensei}
Debrup Das, Debopriyo Banerjee, Somak Aditya, and Ashish Kulkarni. 2024.
\newblock Mathsensei: A tool-augmented large language model for mathematical reasoning.
\newblock \emph{arXiv preprint arXiv:2402.17231}.

\bibitem[{De~Moura and Bj{\o}rner(2008)}]{de2008z3}
Leonardo De~Moura and Nikolaj Bj{\o}rner. 2008.
\newblock Z3: An efficient smt solver.
\newblock In \emph{International conference on Tools and Algorithms for the Construction and Analysis of Systems}, pages 337--340. Springer.

\bibitem[{Genesereth and Kao(2022)}]{genesereth2022introduction}
Michael Genesereth and Eric Kao. 2022.
\newblock \emph{Introduction to logic}.
\newblock Springer Nature.

\bibitem[{Gou et~al.(2023)Gou, Shao, Gong, Shen, Yang, Huang, Duan, and Chen}]{gou2023tora}
Zhibin Gou, Zhihong Shao, Yeyun Gong, Yelong Shen, Yujiu Yang, Minlie Huang, Nan Duan, and Weizhu Chen. 2023.
\newblock Tora: A tool-integrated reasoning agent for mathematical problem solving.
\newblock \emph{arXiv preprint arXiv:2309.17452}.

\bibitem[{Guo et~al.(2025)Guo, Yang, Zhang, Song, Zhang, Xu, Zhu, Ma, Wang, Bi et~al.}]{guo2025deepseek}
Daya Guo, Dejian Yang, Haowei Zhang, Junxiao Song, Ruoyu Zhang, Runxin Xu, Qihao Zhu, Shirong Ma, Peiyi Wang, Xiao Bi, et~al. 2025.
\newblock Deepseek-r1: Incentivizing reasoning capability in llms via reinforcement learning.
\newblock \emph{arXiv preprint arXiv:2501.12948}.

\bibitem[{Hendrycks et~al.(2021)Hendrycks, Burns, Kadavath, Arora, Basart, Tang, Song, and Steinhardt}]{hendrycks2021measuring}
Dan Hendrycks, Collin Burns, Saurav Kadavath, Akul Arora, Steven Basart, Eric Tang, Dawn Song, and Jacob Steinhardt. 2021.
\newblock Measuring mathematical problem solving with the math dataset.
\newblock \emph{arXiv preprint arXiv:2103.03874}.

\bibitem[{Huet et~al.(1997)Huet, Kahn, and Paulin-Mohring}]{huet1997coq}
G{\'e}rard Huet, Gilles Kahn, and Christine Paulin-Mohring. 1997.
\newblock The coq proof assistant a tutorial.
\newblock \emph{Rapport Technique}, 178.

\bibitem[{Kamoi et~al.(2024)Kamoi, Zhang, Zhang, Han, and Zhang}]{kamoi-etal-2024-llms}
Ryo Kamoi, Yusen Zhang, Nan Zhang, Jiawei Han, and Rui Zhang. 2024.
\newblock \href {https://doi.org/10.1162/tacl_a_00713} {When can {LLM}s actually correct their own mistakes? a critical survey of self-correction of {LLM}s}.
\newblock \emph{Transactions of the Association for Computational Linguistics}, 12:1417--1440.

\bibitem[{Khalifa et~al.(2023)Khalifa, Logeswaran, Lee, Lee, and Wang}]{khalifa2023grace}
Muhammad Khalifa, Lajanugen Logeswaran, Moontae Lee, Honglak Lee, and Lu~Wang. 2023.
\newblock Grace: Discriminator-guided chain-of-thought reasoning.
\newblock In \emph{Findings of the Association for Computational Linguistics: EMNLP 2023}, pages 15299--15328.

\bibitem[{Lightman et~al.(2023)Lightman, Kosaraju, Burda, Edwards, Baker, Lee, Leike, Schulman, Sutskever, and Cobbe}]{lightman2023let}
Hunter Lightman, Vineet Kosaraju, Yura Burda, Harri Edwards, Bowen Baker, Teddy Lee, Jan Leike, John Schulman, Ilya Sutskever, and Karl Cobbe. 2023.
\newblock Let's verify step by step.
\newblock \emph{arXiv preprint arXiv:2305.20050}.

\bibitem[{Lin et~al.(2024)Lin, Gou, Liang, Luo, Liu, and Yang}]{lin2024criticbench}
Zicheng Lin, Zhibin Gou, Tian Liang, Ruilin Luo, Haowei Liu, and Yujiu Yang. 2024.
\newblock Criticbench: Benchmarking llms for critique-correct reasoning.
\newblock \emph{arXiv preprint arXiv:2402.14809}.

\bibitem[{Ling et~al.(2024)Ling, Fang, Li, Huang, Lee, Memisevic, and Su}]{ling2024deductive}
Zhan Ling, Yunhao Fang, Xuanlin Li, Zhiao Huang, Mingu Lee, Roland Memisevic, and Hao Su. 2024.
\newblock Deductive verification of chain-of-thought reasoning.
\newblock \emph{Advances in Neural Information Processing Systems}, 36.

\bibitem[{Liu et~al.(2024{\natexlab{a}})Liu, Feng, Xue, Wang, Wu, Lu, Zhao, Deng, Zhang, Ruan et~al.}]{liu2024deepseek}
Aixin Liu, Bei Feng, Bing Xue, Bingxuan Wang, Bochao Wu, Chengda Lu, Chenggang Zhao, Chengqi Deng, Chenyu Zhang, Chong Ruan, et~al. 2024{\natexlab{a}}.
\newblock Deepseek-v3 technical report.
\newblock \emph{arXiv preprint arXiv:2412.19437}.

\bibitem[{Liu et~al.(2024{\natexlab{b}})Liu, Sun, Li, and Yao}]{liu2024efficient}
Haoxiong Liu, Jiacheng Sun, Zhenguo Li, and Andrew~C Yao. 2024{\natexlab{b}}.
\newblock Efficient neural theorem proving via fine-grained proof structure analysis.
\newblock \emph{arXiv preprint arXiv:2501.18310}.

\bibitem[{Luo et~al.(2023)Luo, Lin, Liu, Shu, Zhu, Shang, and Meng}]{luo2023critique}
Liangchen Luo, Zi~Lin, Yinxiao Liu, Lei Shu, Yun Zhu, Jingbo Shang, and Lei Meng. 2023.
\newblock Critique ability of large language models.
\newblock \emph{arXiv preprint arXiv:2310.04815}.

\bibitem[{Madaan et~al.(2024)Madaan, Tandon, Gupta, Hallinan, Gao, Wiegreffe, Alon, Dziri, Prabhumoye, Yang et~al.}]{madaan2024self}
Aman Madaan, Niket Tandon, Prakhar Gupta, Skyler Hallinan, Luyu Gao, Sarah Wiegreffe, Uri Alon, Nouha Dziri, Shrimai Prabhumoye, Yiming Yang, et~al. 2024.
\newblock Self-refine: Iterative refinement with self-feedback.
\newblock \emph{Advances in Neural Information Processing Systems}, 36.

\bibitem[{Meurer et~al.(2017)Meurer, Smith, Paprocki, {\v{C}}ert{\'\i}k, Kirpichev, Rocklin, Kumar, Ivanov, Moore, Singh et~al.}]{meurer2017SymPy}
Aaron Meurer, Christopher~P Smith, Mateusz Paprocki, Ond{\v{r}}ej {\v{C}}ert{\'\i}k, Sergey~B Kirpichev, Matthew Rocklin, AMiT Kumar, Sergiu Ivanov, Jason~K Moore, Sartaj Singh, et~al. 2017.
\newblock Sympy: symbolic computing in python.
\newblock \emph{PeerJ Computer Science}, 3:e103.

\bibitem[{Mirzadeh et~al.(2024)Mirzadeh, Alizadeh, Shahrokhi, Tuzel, Bengio, and Farajtabar}]{mirzadeh2024gsm}
Iman Mirzadeh, Keivan Alizadeh, Hooman Shahrokhi, Oncel Tuzel, Samy Bengio, and Mehrdad Farajtabar. 2024.
\newblock Gsm-symbolic: Understanding the limitations of mathematical reasoning in large language models.
\newblock \emph{arXiv preprint arXiv:2410.05229}.

\bibitem[{Moura and Ullrich(2021)}]{moura2021lean}
Leonardo~de Moura and Sebastian Ullrich. 2021.
\newblock The lean 4 theorem prover and programming language.
\newblock In \emph{Automated Deduction--CADE 28: 28th International Conference on Automated Deduction, Virtual Event, July 12--15, 2021, Proceedings 28}, pages 625--635. Springer.

\bibitem[{Olausson et~al.(2023)Olausson, Gu, Lipkin, Zhang, Solar-Lezama, Tenenbaum, and Levy}]{olausson2023linc}
Theo~X Olausson, Alex Gu, Ben Lipkin, Cedegao~E Zhang, Armando Solar-Lezama, Joshua~B Tenenbaum, and Roger~P Levy. 2023.
\newblock Linc: A neurosymbolic approach for logical reasoning by combining language models with first-order logic provers.
\newblock In \emph{The 2023 Conference on Empirical Methods in Natural Language Processing}.

\bibitem[{Pan et~al.(2023)Pan, Albalak, Wang, and Wang}]{pan2023logic}
Liangming Pan, Alon Albalak, Xinyi Wang, and William~Yang Wang. 2023.
\newblock Logic-lm: Empowering large language models with symbolic solvers for faithful logical reasoning.
\newblock In \emph{The 2023 Conference on Empirical Methods in Natural Language Processing}.

\bibitem[{Parnami and Lee(2022)}]{parnami2022learning}
Archit Parnami and Minwoo Lee. 2022.
\newblock Learning from few examples: A summary of approaches to few-shot learning.
\newblock \emph{arXiv preprint arXiv:2203.04291}.

\bibitem[{Pathak(2024)}]{pathak2024gflean}
Shashank Pathak. 2024.
\newblock Gflean: An autoformalisation framework for lean via gf.
\newblock \emph{arXiv preprint arXiv:2404.01234}.

\bibitem[{Ranta(2004)}]{ranta2004grammatical}
Aarne Ranta. 2004.
\newblock Grammatical framework.
\newblock \emph{Journal of Functional Programming}, 14(2):145--189.

\bibitem[{Raza and Milic-Frayling(2025)}]{raza2025instantiation}
Mohammad Raza and Natasa Milic-Frayling. 2025.
\newblock Instantiation-based formalization of logical reasoning tasks using language models and logical solvers.
\newblock \emph{arXiv preprint arXiv:2501.16961}.

\bibitem[{Schaefer and Kohlhase(2020)}]{schaefer2020glif}
Jan~Frederik Schaefer and Michael Kohlhase. 2020.
\newblock Glif: A declarative framework for symbolic natural language understanding.
\newblock In \emph{FCR@ KI}, pages 4--11.

\bibitem[{Song et~al.(2025)Song, Su, Qu, Zhou, and Cheng}]{song2025prmbench}
Mingyang Song, Zhaochen Su, Xiaoye Qu, Jiawei Zhou, and Yu~Cheng. 2025.
\newblock Prmbench: A fine-grained and challenging benchmark for process-level reward models.
\newblock \emph{arXiv preprint arXiv:2501.03124}.

\bibitem[{Sun et~al.(2025)Sun, Yin, Huang, Qiu, and Zhao}]{sun2025error}
Yuhong Sun, Zhangyue Yin, Xuanjing Huang, Xipeng Qiu, and Hui Zhao. 2025.
\newblock Error classification of large language models on math word problems: A dynamically adaptive framework.
\newblock \emph{arXiv preprint arXiv:2501.15581}.

\bibitem[{{\'S}wiechowski et~al.(2023){\'S}wiechowski, Godlewski, Sawicki, and Ma{\'n}dziuk}]{swiechowski2023monte}
Maciej {\'S}wiechowski, Konrad Godlewski, Bartosz Sawicki, and Jacek Ma{\'n}dziuk. 2023.
\newblock Monte carlo tree search: A review of recent modifications and applications.
\newblock \emph{Artificial Intelligence Review}, 56(3):2497--2562.

\bibitem[{Toh et~al.(2024)Toh, Ghosal, and Poria}]{toh2024votescountprogramsverifiers}
Vernon Y.~H. Toh, Deepanway Ghosal, and Soujanya Poria. 2024.
\newblock \href {https://arxiv.org/abs/2410.12608} {Not all votes count! programs as verifiers improve self-consistency of language models for math reasoning}.
\newblock \emph{Preprint}, arXiv:2410.12608.

\bibitem[{Wang et~al.(2024)Wang, Li, Shao, Xu, Dai, Li, Chen, Wu, and Sui}]{wang-etal-2024-math}
Peiyi Wang, Lei Li, Zhihong Shao, Runxin Xu, Damai Dai, Yifei Li, Deli Chen, Yu~Wu, and Zhifang Sui. 2024.
\newblock \href {https://doi.org/10.18653/v1/2024.acl-long.510} {Math-shepherd: Verify and reinforce {LLM}s step-by-step without human annotations}.
\newblock In \emph{Proceedings of the 62nd Annual Meeting of the Association for Computational Linguistics (Volume 1: Long Papers)}, pages 9426--9439, Bangkok, Thailand. Association for Computational Linguistics.

\bibitem[{Wang et~al.()Wang, Wei, Schuurmans, Le, Chi, Narang, Chowdhery, and Zhou}]{wangself}
Xuezhi Wang, Jason Wei, Dale Schuurmans, Quoc~V Le, Ed~H Chi, Sharan Narang, Aakanksha Chowdhery, and Denny Zhou.
\newblock Self-consistency improves chain of thought reasoning in language models.
\newblock In \emph{The Eleventh International Conference on Learning Representations}.

\bibitem[{Wang et~al.(2020)Wang, Yao, Kwok, and Ni}]{wang2020generalizing}
Yaqing Wang, Quanming Yao, James~T Kwok, and Lionel~M Ni. 2020.
\newblock Generalizing from a few examples: A survey on few-shot learning.
\newblock \emph{ACM computing surveys (csur)}, 53(3):1--34.

\bibitem[{Wu et~al.(2022)Wu, Jiang, Li, Rabe, Staats, Jamnik, and Szegedy}]{wu2022autoformalization}
Yuhuai Wu, Albert~Qiaochu Jiang, Wenda Li, Markus Rabe, Charles Staats, Mateja Jamnik, and Christian Szegedy. 2022.
\newblock Autoformalization with large language models.
\newblock \emph{Advances in Neural Information Processing Systems}, 35:32353--32368.

\bibitem[{Xi et~al.(2024)Xi, Yang, Huang, Tang, Li, Ding, He, Hong, Do, Zhan et~al.}]{xi2024enhancing}
Zhiheng Xi, Dingwen Yang, Jixuan Huang, Jiafu Tang, Guanyu Li, Yiwen Ding, Wei He, Boyang Hong, Shihan Do, Wenyu Zhan, et~al. 2024.
\newblock Enhancing llm reasoning via critique models with test-time and training-time supervision.
\newblock \emph{arXiv preprint arXiv:2411.16579}.

\bibitem[{Xin et~al.(2024{\natexlab{a}})Xin, Guo, Shao, Ren, Zhu, Liu, Ruan, Li, and Liang}]{xin2024deepseekv1}
Huajian Xin, Daya Guo, Zhihong Shao, Zhizhou Ren, Qihao Zhu, Bo~Liu, Chong Ruan, Wenda Li, and Xiaodan Liang. 2024{\natexlab{a}}.
\newblock Deepseek-prover: Advancing theorem proving in llms through large-scale synthetic data.
\newblock \emph{arXiv preprint arXiv:2405.14333}.

\bibitem[{Xin et~al.(2024{\natexlab{b}})Xin, Ren, Song, Shao, Zhao, Wang, Liu, Zhang, Lu, Du et~al.}]{xin2024deepseekv1_5}
Huajian Xin, ZZ~Ren, Junxiao Song, Zhihong Shao, Wanjia Zhao, Haocheng Wang, Bo~Liu, Liyue Zhang, Xuan Lu, Qiushi Du, et~al. 2024{\natexlab{b}}.
\newblock Deepseek-prover-v1. 5: Harnessing proof assistant feedback for reinforcement learning and monte-carlo tree search.
\newblock \emph{arXiv preprint arXiv:2408.08152}.

\bibitem[{Ye et~al.(2024)Ye, Chen, Dillig, and Durrett}]{ye2024satlm}
Xi~Ye, Qiaochu Chen, Isil Dillig, and Greg Durrett. 2024.
\newblock Satlm: Satisfiability-aided language models using declarative prompting.
\newblock \emph{Advances in Neural Information Processing Systems}, 36.

\bibitem[{Zhang et~al.(2025)Zhang, Zheng, Wu, Zhang, Lin, Yu, Liu, Zhou, and Lin}]{zhang2025lessons}
Zhenru Zhang, Chujie Zheng, Yangzhen Wu, Beichen Zhang, Runji Lin, Bowen Yu, Dayiheng Liu, Jingren Zhou, and Junyang Lin. 2025.
\newblock The lessons of developing process reward models in mathematical reasoning.
\newblock \emph{arXiv preprint arXiv:2501.07301}.

\bibitem[{Zheng et~al.(2024)Zheng, Zhang, Zhang, Lin, Lu, Yu, Liu, Zhou, and Lin}]{zheng2024processbench}
Chujie Zheng, Zhenru Zhang, Beichen Zhang, Runji Lin, Keming Lu, Bowen Yu, Dayiheng Liu, Jingren Zhou, and Junyang Lin. 2024.
\newblock Processbench: Identifying process errors in mathematical reasoning.
\newblock \emph{arXiv preprint arXiv:2412.06559}.

\bibitem[{Zhou et~al.(2024{\natexlab{a}})Zhou, Staats, Li, Szegedy, Weinberger, and Wu}]{zhoudon}
Jin~Peng Zhou, Charles~E Staats, Wenda Li, Christian Szegedy, Kilian~Q Weinberger, and Yuhuai Wu. 2024{\natexlab{a}}.
\newblock Don't trust: Verify--grounding llm quantitative reasoning with autoformalization.
\newblock In \emph{The Twelfth International Conference on Learning Representations}.

\bibitem[{Zhou et~al.(2024{\natexlab{b}})Zhou, Schellaert, Mart{\'\i}nez-Plumed, Moros-Daval, Ferri, and Hern{\'a}ndez-Orallo}]{zhou2024larger}
Lexin Zhou, Wout Schellaert, Fernando Mart{\'\i}nez-Plumed, Yael Moros-Daval, C{\`e}sar Ferri, and Jos{\'e} Hern{\'a}ndez-Orallo. 2024{\natexlab{b}}.
\newblock Larger and more instructable language models become less reliable.
\newblock \emph{Nature}, 634(8032):61--68.

\end{thebibliography}

\newpage

\appendix

\newpage

\section{Example of Critic Workflow}
\label{sec:appendix1}
\begin{figure*}[htpb]
 \centering
  \includegraphics[width=\linewidth]{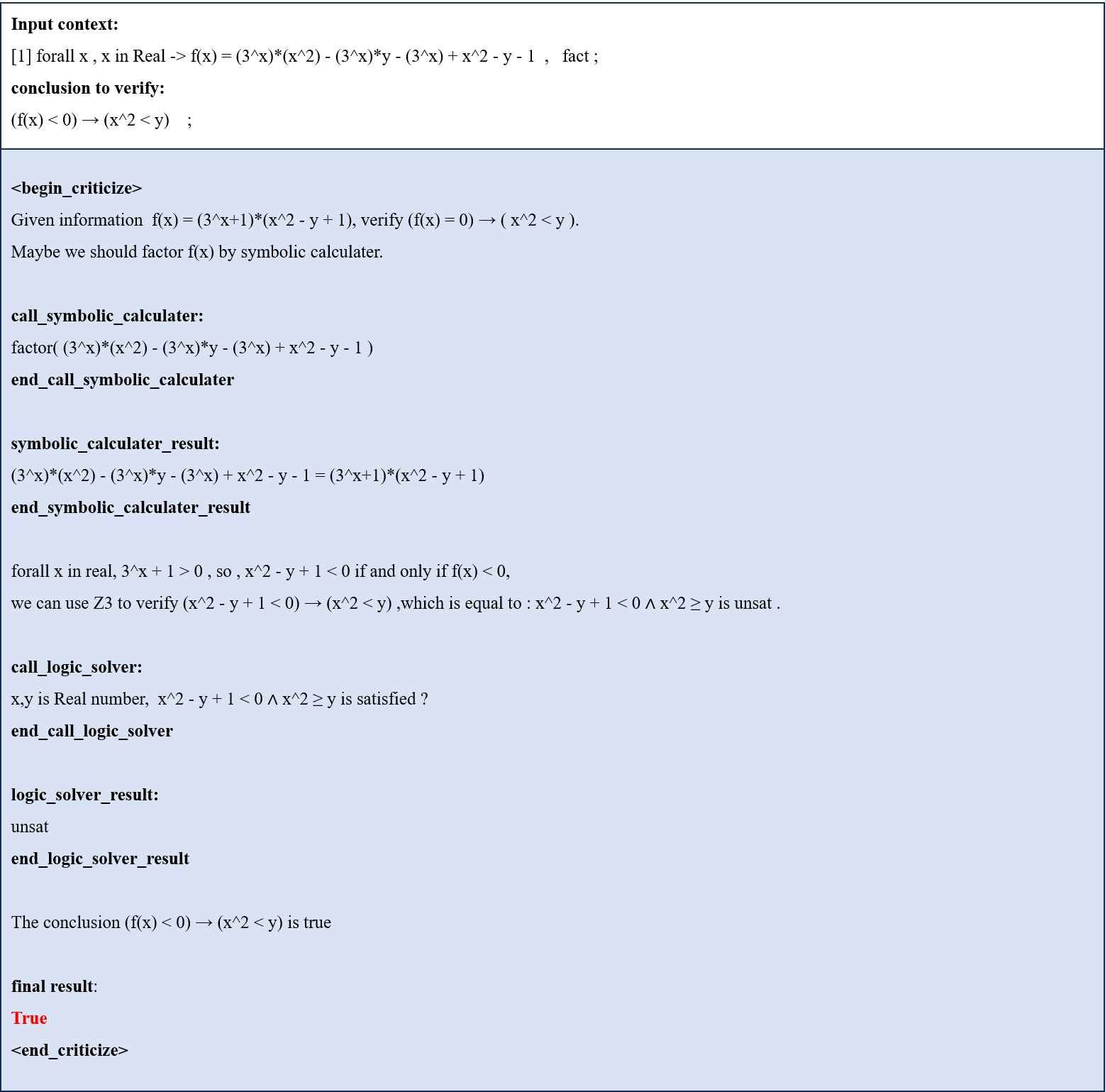}
  \caption{The example of critic workflow. In this example, we input context and the conclusion to verify, and then our Critic first call symbolic calculater and then call logic solver, subsequently concluding that this judgment is true.}
  % \caption{Overview of MATH-VF: First, the problem and solution are input into the formalizer, resulting in a context of the formal solution. And then We decompose the issue of correctness for the formal solution into a series of correctness issues for individual judgments.}
  \label{fig:example1}
\end{figure*}

\end{document}